\documentclass{article}

\usepackage[utf8]{inputenc} 
\usepackage[T1]{fontenc}    
\usepackage{hyperref}       
\usepackage{url}            
\usepackage{booktabs}       
\usepackage{amsfonts}       
\usepackage{nicefrac}       
\usepackage{microtype}      
\usepackage{graphicx}
\usepackage{doi}
\usepackage{natbib}

\title{How Can Information Behaviour Inform Machine Learning?}

\author{Michael Ridley \\
	Librarian Emeritus\\
	University of Guelph\\
	Guelph, Ontario, Canada N1L 2W1 \\
	\texttt{mridley@uoguelph.ca} \\
}

\hypersetup{
pdftitle={How Can Information Behaviour Inform Machine Learning?},
pdfauthor={Michael ~Ridley},
pdfkeywords={information behaviour, machine learning, applied research},
}

\begin{document}
\maketitle

\begin{abstract}
 The objective of this paper is to explore the opportunities for human information behaviour research to inform and influence the field of machine learning and the resulting machine information behaviour. Using the development of foundation models in machine learning as an example, the paper illustrates how human information behaviour research can bring to machine learning a more nuanced view of information and informing, a better understanding of information need and how that affects the communication among people and systems, guidance on the nature of context and how to operationalize that in models and systems, and insights into bias, misinformation, and marginalization. Despite their clear differences, the fields of information behaviour and machine learning share many common objectives, paradigms, and key research questions. The example of foundation models illustrates that human information behaviour research has much to offer in addressing some of the challenges emerging in the nascent area of machine information behaviour.
\end{abstract}

\section{Introduction}
Advances in machine learning advances have led to the emergent field of machine information behaviour where algorithmic systems exhibit information seeking and use characteristics that are analogous to human information behaviour. However, machine information behaviour is nascent, and the machine learning community has yet to leverage what we know about human information behaviour to understand and enhance machine information behaviour. Hence, the key question, how can information behaviour inform machine learning?

In 1999 Marcia Bates warned that

\begin{quote}
    the wheel is being reinvented every day on the information superhighway. Our expertise is ignored while newcomers to information questions stumble through tens of millions of dollars of research and startup money to rediscover what information science knew in 1960. We in the field need to make our research and theory better known and more understandable to the newcomers flooding in—or be washed away in the flood. (Bates, 1999, p. 1043)
\end{quote}

Over 20 years later, two Google staff researchers presented a paper at the annual conference of the Association for the Advancement of Artificial Intelligence entitled "Model-agnostic fits for understanding information seeking in humans" (Chatterjee and Shenoy, 2021). While a good paper, it includes, perhaps unsurprising to Bates, not a single reference or allusion to the extensive work on information seeking from the field of information behaviour. This is not an isolated instance. 

In 1854 Thoreau famously wrote, "We are in great haste to construct a magnetic telegraph from Maine to Texas, but Maine and Texas, it may be, have nothing important to communicate" (Thoreau, 1854/2014, p. 50). If the analysis presented here is compelling, not only do information behaviour and machine learning have much to communicate but those conversations are more important than ever.

The fields of information behaviour and machine learning share many common objectives, paradigms, and key research questions. Machine learning is deeply and fundamentally engaged with information. It is also a field that has repeatedly looked to human information processing as a source for inspiration and computational analogs. Neural networks (i.e., "artificial" neural networks), the basis for much machine learning, are the most prominent example of this (LeCun et al., 2015). Machine learning and information behaviour also share a focus on learning. It is explicit in machine learning and implicated in information behaviour (Case and Given, 2016; Wilson, 2016). As a result, these fields share a "socio-techno-informational" nexus (Eriksson-Backa et al., 2021; Huvila et al., 2021) that offers the basis for exchange.

Despite this complementarity, machine learning and information behaviour largely operate as two solitudes. Foundation models, such as GPT-3 from OpenAI (Tamkin et al., 2021), represent a major advancement in machine learning. By examining foundation models through the lens of information behaviour, the opportunities for information behaviour research to inform and influence machine learning are explored.

\section{Structure of the Paper}
This paper first reviews the limited intersection of information behaviour and machine learning research. Foundation models are then defined and described, including issues and challenges raised by developers and critics. Following this, insights and findings from information behaviour research are applied to those challenges. As part of this discussion, the emerging attention to neurosymbolic machine learning is presented as a new opportunity to apply information behaviour research. The paper concludes by highlighting the importance of applied information behaviour and suggests ways to bridge the divide between information behaviour and machine learning.

\section{Machine Learning and Information Behaviour Research}
Information behaviour research has been adopted by other disciplines including computer science (Huvila et al., 2021; Makri, 2020; Wilson, 2018, 2020b), but it has not influenced the subfield of machine learning. A search for papers discussing topics such as "information behaviour", "information-seeking behaviour", "information-seeking", or "information needs" in the Digital Library of the Association for Computing Machinery (\href{https://dl.acm.org)}{dl.acm.org} and the arXiv pre-print repository (\href{https://arxiv.org}{arXiv.org}) returned few correlations with machine learning research and none with references to information behaviour research. Conversely, while there has been within library and information science an increased interest in data science, artificial intelligence, and specifically algorithms, this trend is not apparent in the information behaviour field (Ma and Lund, 2021). 

As Julien and O’Brien note, for information behaviour research ‘to have a greater impact beyond the borders of information science’ it must, among other things, ‘explicitly articulate the ways in which findings have value for information systems’ (Julien and O’Brien, 2014, p. 248). Seven years later the call for the ‘actionable implications’ of information behaviour research is repeated, highlighting both the real and imagined ‘gap’ between information behavior research and information systems design (Huvila et al., 2021, p. 6). The most prominent and important contemporary information systems are those based on machine learning.

The field of information behaviour examines how ‘people need, seek, manage, give, and use information in different contexts’ (Fisher et al., 2005, p. xix). The concept of machine behaviour, encourages researchers to view machine learning systems (i.e., intelligent systems) through the lens of behavioural concepts (Rahwan et al., 2019). Following this, it is possible to define machine information behaviour as how intelligent machines need, seek, manage, give and use information in different contexts (Ridley, 2019, 2022). It is the objective of this paper to illustrate that human information behaviour can inform the machine information behaviour created through machine learning.

In recent years the machine learning community has acknowledged the need for multidisciplinary perspectives and engagement (Littman et al., 2021). The limitations of prior research and development have led to a focus on human-centered artificial intelligence where intelligent systems must ‘understand the (often culturally specific) expectations and needs of humans and to help humans understand them in return’ (Riedl, 2019, p. 36). The unexplored alignment between information behaviour and machine learning presents an opportunity for information behaviour research to inform a field eager for new perspectives.

\section{Foundation Models}
A foundation model ‘is any model that is trained on broad data at scale and can be adapted (e.g., fine-tuned) to a wide range of downstream tasks’ (Bommasani et al., 2021, p. 3). Foundation models, typified by models such as BERT and GPT-3, are trained on petabytes of data, contain billion of parameters, and use deep neural networks and self-supervised learning methods (Brown et al., 2020). These models ‘learn’ the structure and semantics of language at scale and apply that to the information from the domain specific training data. Foundation models mark a significant advance in both natural language understanding and natural language processing.

A foundation model can be adapted (e.g., augmented or fine-tuned) with task or domain specific data to apply successfully to different contexts. As a result, one foundation model, such as GPT-3, can spawn many other models used for specific purposes often well beyond the original objective. For example, GPT-3 was trained on large language corpora and with a few adaptions was able to generate functional computer code (Chen et al., 2021). Initially limited to textual training data, emerging foundation models and their derivatives are increasingly multimodal (e.g., text, sound, image) (Bommasani et al., 2021; Yuan et al., 2021).

Foundation models are applied in fields such organic compounds (Rothchild et al., 2021) and climate change (Lacoste et al., 2021), with further applications expected in health care with diagnostic interfaces for health care providers and question/answering systems for patients, both trained on a vast amount of biomedical data (Bommasani et al., 2021). Google is using foundation models, Multitask Unified Model (MUM) and LaMDA, in its search application (Collins and Ghahramani, 2021; Nayak, 2021). Rather than retrieving pages or documents, a search results in a machine learning generated textual response (e.g., an answer, recommendation, or description) and initiates a conversation should the user wish to explore further.

In summary, the key objectives of foundation models are to ‘(1) distill and accumulate knowledge from various sources and domains, (2) organize it in an effective and scalable representation, and (3) flexibly generalize it towards novel contexts’ (Bommasani et al., 2021, p. 74). Or, to use the terminology of information behaviour, to seek, manage, and use information in context. These objectives for foundation models have resulted in ‘unprecedented challenges for understanding their behavior’ (Bommasani et al., 2021, p. 122). As with neural networks previously, human information behaviour provides a source of inspiration and computational analogs to respond to those challenges.

\section{Foundation Models: Issues and Challenges}
Foundation models have been criticized as “stochastic parrots” (Bender et al., 2021) and “parlor tricks” (Marcus and Davis, 2021) that “often fail to generalize robustly and are susceptible to spurious regularities in the training data” (Russell, 2021, p. 512). As such they are a source of significant societal risk (Bommasani et al., 2021; Tamkin et al., 2021; Weidinger et al., 2021). 

Despite the deficiencies, gaffs, and worse, foundation systems have proven exceptionally powerful and successful. They have surpassed and supplanted many of the techniques previously central to machine learning:

\begin{quote}
    The problem is that they [foundation models] got good. Really good. And the rest of our natural language toolkit, the one that connects these models to the real world, hasn’t kept pace. Knowledge graphs as a source of verified facts haven’t kept pace. Multimodal learning as a source of grounding hasn’t kept pace. Common sense reasoning hasn’t kept pace. Our grand challenge today is to rectify this: find new, scalable ways to anchor these language models in reality, imbue them with common sense and logic, and ensure that their worldview is grounded in facts instead of internet memes. (Vanhoucke, 2021)
\end{quote}

The challenges of anchoring, imbuing, and grounding in foundation models can be viewed in the context of their two key characteristics, emergence, and homogenization: 

\begin{quote}
    Emergence means that the behavior of a system is implicitly induced rather than explicitly constructed; it is both the source of scientific excitement and anxiety about unanticipated consequences. Homogenization indicates the consolidation of methodologies for building machine learning systems across a wide range of applications; it provides strong leverage towards many tasks but also creates single points of failure. (Bommasani et al., 2021, p. 3)
\end{quote}

Emergent behavior, arising from self-supervised learning, creates challenges regarding generalization, contextuality, and compositionality, and exposes the models to bias, harms, and the proliferation of misinformation and disinformation. Those challenges can result from the training data, the representation of that data, and the biases of data modelers and system developers. Homogenization leverages common techniques across many domains (especially important for those where task specific data is limited). However, this also moves towards “a single generic learning algorithm” (Bommasani et al., 2021, p. 4) which reduces diversity in the algorithmic processes and introduces “inertia” (Steinhardt, 2021) that impedes the identification of problems and delays correction (i.e., retraining of the model).

Deficiencies in the foundation model will compromise the derived models, and derived models themselves introduce their own challenges. Most people will experience foundation models through the task and domain specific derived models (e.g., Google search). The processes of augmenting a foundation model through temporal adaptation (updating information and inferences), domain specialization (including specific training data), and the application of constraints (e.g., to protect privacy) inherit and amplify in the derived model the latent problems of the originating foundation model.

\section{Lessons from Information Behaviour}
The ten lessons of information behaviour (Case and Given, 2016), derived and expanded from Dervin’s ten assumptions (i.e., myths) (Dervin, 1976) are a useful starting point. A sample of these, with the research that references them, offers key insights into how information behaviour can inform machine learning. To be clear, this analysis is not intended as a comprehensive review of information behaviour research but an illustrative, and necessarily selective, reading of the literature.

\subsection{"Formal sources and rationalized searches reflect only one side of human information behavior"}
Recorded information and intentional information seeking are the hallmarks of foundation systems. These perspectives, with respect to the training data and the user interactions, focus on “the ‘information’ and not the ‘informing’” (Dervin, 1976, p. 328). The information experience, what is and what is not informative, positions information as a process as much as it is objective thing (Gorichanaz, 2020). Information behaviour offers a more nuanced view of information seeking that acknowledges less formal and more passive modes of information discovery (Wilson, 2020a) and information encountering (Erdelez, 1997). 

Foundation systems ingest recorded information and ignore the many non-documentary means of information behaviour, such as conferring with colleagues or friends (Hanlon and McLeod, 2020), reinforcing another lesson: “Better system design will not eliminate the need for interpersonal communication. The consistent finding across information behaviour research on sources, channels, and types of users is that ‘people still turn to other people for information’” (Case and Given, 2016, p. 346). The social dimension of information needs and information seeking are absent from foundation models.

While the simulated conversations of foundation models will improve and prove valuable, the absence of a social dimension will still undermine the objective of human-centred artificial intelligence. The distinction is the difference between traditional human-computer interaction and human-machine communication where the focus of the latter is on “people’s interactions with technologies designed as communicative subjects, instead of mere interactive objects ... machine as communicator” (Guzman and Lewis, 2020, p. 71). The one-to-one of human-computer interaction is now a many-to-many communication environment where human and computer agents intermingle. It is not a matter of focusing on either users or systems but applying information behavior to the integrated sociotechnical assemblages arising from intelligent systems.

Theories of information need (Belkin, 1980; Dervin, 1976; Kuhlthau, 1991; R. S. Taylor, 1968) introduce notions of uncertainty, gaps, and sensemaking that motivate users. Information need is understood not as an isolated and singular condition, but something that manifests itself over time (Cole, 2012). Foundation systems and their derivatives view information need in a transactional manner (e.g., question and answer; statements and responses). The nuances of need, central to human information behaviour, are unexplored and replaced with superficial assumptions about the user.

\subsection{“Context is central to the transfer of information”}

That context matters is perhaps the central tenet of information behaviour (Agarwal, 2017; Case and Given, 2016). While recognized as a challenge for foundation models, the implications of context are poorly understood and as a result poorly operationalized in these systems. Context in foundation models is typically conceptualized as adaptation: how a system responds to a changed task or domain. This limited view of context prevents foundation models and their derivatives from leveraging the advantage of context in using information and from differentiating the motivations of users of the system.

Context in machine learning, as with most things in this field, is a mathematical concept. Vectorizing the training data is a process that creates a complex, multidimensional information space but also a flatness in the data thereby losing critical, contextual information. Relationships in vectors are determined by “distance” measures such as cosine similarity and Euclidean distance. Advances using transformer and attention techniques to assess semantic context are, by information behaviour standards, blunt instruments.

With information mostly decontextualized through vectorization, considerable information is stripped regarding authorship, publication, and characteristics concerning time and space (the latter acknowledging Dervin’s myth that “Time and space can be ignored”). This is a further reminder both that “raw data is an oxymoron” (Gitelman, 2013) and that any data representation or classification has consequences (Bowker and Star, 1999).

While foundation models are self-supervised learning environments that does not and cannot exclude the interventions of system developers (“modelers”) from imposing conditions and constraints on the way the data is presented to the algorithms. Developers bring their biases and assumptions to their work. Despite the opacity of machine learning systems, the influence of system developers is a reminder that “the ‘black box’ is full of people” (Seaver, 2021, p. 773).

Data in foundational systems consists of extensive data sources used to train the agnostic language model followed by task or domain data to train for specific purposes. In both cases the nature of the training data is critical. Many machine learning systems use benchmark datasets, such as ImageNet (http://image-net.org/) and CIFAR-10 (http://www.cs.toronto.edu/~kriz/cifar.html), to evaluate models. Deficiencies in these datasets has led to calls for synthetic benchmark datasets with purpose-built challenges and objectives to better evaluate the models (Nikolenko, 2019; Paullada et al., 2020). There are opportunities for information behaviour researchers to contribute to the design and content of these synthetic datasets to test models for relevant issues such as contextual understanding, detection of bias and misinformation, issues regarding data representation, and frameworks of user expectations and behaviour.

\subsection{“More information is not always better”}

Machine learning is dominated by Dervin’s myth version of this lesson: “If a little information is good, a lot must be better.” Ingesting more data is often the machine learning solution to overfitting or underfitting models (Burkov, 2019). Foundation models are fundamentally about data at scale with training done on petabytes of data. Given the self-supervised learning at the heart of foundation models, more data can reinforce existing biases and misinformation (Tamkin et al., 2021). 

Foundation models are clearly successful in the objective to accumulate vast amounts of information. However, the objective to distill that information in ways that identify and resolve bias and misinformation has been less successful. As developers of foundation models acknowledge “the underlying statistical methods are not well-positioned to distinguish between factually correct and incorrect information” (Weidinger et al., 2021, p. 1). Information behaviour research has long been interested in misinformation, developing both useful models to understand it (Karlova and Fisher, 2013) and recommendations on how to mitigate it (Rubin, 2019). More broadly, the antidote to simply more information may be an “explicit focus on the usefulness of information versus the usefulness of IS [information systems]” (Huvila et al., 2021, p. 10).

Bias and unfairness have become a focus of machine learning research with specific concerns regarding foundation models. A prominent annual conference is now devoted to this work: the Association of Computing Machinery Conference on Fairness, Accountability, and Transparency (http://facctconference.org). With respect to bias and unfairness, information behaviour research on marginalized communities has much to offer. Extending Chatman’s description of information poverty (Chatman, 1996), information marginalization is defined as “the institutional and or community-level mechanisms by which information poverty is created” (Gibson and Martin, 2019, p. 477). Those involved in machine learning research and development represent one type of those institutions and communities. Foundation models and other machine learning systems are built and trained within an academic and professional milieu often motivated by financial objectives. While claiming to be neutral, machine learning has been found to be “socially and politically loaded, frequently neglecting societal needs and harms, while prioritizing and promoting the concentration of power in the hands of already powerful actors” (Birhane et al., 2021, p. 10). One result of this is the “systemic failure of information systems to meet the needs of marginalized groups of people” (Gibson and Martin, 2019, p. 486). Some of the most prominent and influential information behaviour research has identified the marginalization of different groups and points to remedies that machine learning can implement (Willson et al., 2021).

Concerns about bias and misinformation in foundation models have directed attention to inverse reinforcement learning as an alternative to, or an augmentation of, deep learning. Inverse reinforcement learning looks to domain experts to establish the rewards and optimal policies that guide intelligent systems (Zhifei and Meng Joo, 2012). While similar to the expert systems prominent at the end of the 20th century (Hayes-Roth et al., 1983), inverse reinforcement learning research has explored more effective ways to incorporate expert knowledge (M. E. Taylor, 2018). The contribution information behaviour can make is the understanding about how different experts seek, manage, and use information. Insights into the information behaviour of groups such as scientists (Ellis et al., 1993), engineers (Fidel and Green, 2004), managers (Choo and Auster, 1993), and other professionals, (Leckie et al., 1996) provide valuable context for eliciting expertise for inverse reinforcement learning.

\section{Information Behaviour Models}

Information behaviour research has created frameworks and theoretical models that are general as well as contextually specific. Many of these would be relevant as guides for understanding and modeling machine information behaviour through the lens of human information behaviour.

Particularly useful in the context of foundation models are the information behaviour models of everyday information practices, (McKenzie, 2003; Ocepek, 2018; Savolainen, 2008). Since foundation models create information algorithmically through summarization and inference, especially noteworthy is the expansion of Savolainen’s model to encompass information creation (Savolainen and Thomson, 2021). In this elaboration there are two modes of everyday information practices: acquiring and expressing. With information creation and the expressing mode, “attention is turned to a whole host of day-to-day generative information activities” (Savolainen and Thomson, 2021, p. 9). Key to the success of foundation models are their generative and emergent capabilities. The expanded everyday information practices model provides a framework to understand and guide those generative practices.

Another useful model in the context of foundation models is the nonlinear model of information seeking behaviour (Foster, 2004). This model aligns with the nonlinear functions of neural networks and challenges the all too linear user interface assumed by most implementations of foundation models. The core processes (opening, orientation, consolidation) and the cognitive approaches (flexible/adaptive, openness, nomadic, holistic) offer ways that foundation models and their derivatives can rearchitect the underlying algorithms of the neural network and reimagine the user interactions.

Frameworks and models attempt to bring together components and processes into a unified vision. The concept of information behaviour patterns is an attempt to do something similar which may be useful to machine learning. Information behaviour patterns, which identify the relationships among discrete components of information behaviour, are proposed as “beneficial for motivating the design of information environments that promote smooth, less effortful transitions between different types of information acquisition” (Lee et al., 2021, p. 13). This perspective indicates that techniques such as berrypicking (Bates, 1989), browsing (Bates, 1989; Chang and Rice, 1993), optimal foraging (Sandstrom, 1994), triangulation (Greyson, 2018), equilibrium (Liu, 2017), and serendipity (Agarwal, 2015; McCay-Peet and Toms, 2018) are not mutually exclusive but rather an interrelated set of options or strategies. The opportunity for machine learning systems is design for patterns of information acquisition by promoting an ensemble view of techniques that leverage the advantages of each in an appropriate context.

\section{Neurosymbolic Machine Learning: Beyond Foundation Models}

Emerging developments in machine learning may broaden opportunities for the application of information behaviour research. Critics of foundation models propose a neurosymbolic approach that integrates the methods of deep neural networks with symbolic information processing (Marcus, 2020; Marcus and Davis, 2019; Russell, 2019).

The self-supervised learning of foundation models acknowledges Rich Sutton’s observation about the source of machine learning’s success:

\begin{quote}
    The biggest lesson that can be read from 70 years of AI research is that general methods that leverage computation are ultimately the most effective, and by a large margin ... We have to learn the bitter lesson that building in how we think does not work in the long run. (Sutton, 2019)
\end{quote}

As a result, computational capacities (e.g., more and faster CPUs and TPUs, and general-purpose algorithms) are more important and successful than incorporating human information structures, assumptions, and innate knowledge. These factors are what machine learning research calls “priors” and are part of Sutton’s aversion to including “how we think” in machine learning systems. 

However, many feel that machine learning capabilities have plateaued and that foundation models will not by themselves advance the field. The importance of prior knowledge such as knowledge graphs and symbolic information processing tied to causality, contextuality, and adaptation are advanced as ways to augment deep neural networks. The goal is to “integrate deep learning, which excels at perceptual classification, with symbolic systems, which excel at inference and abstraction” (Marcus, 2018, p. 20).

Information behaviour research understands humans as more than mechanistic information processing systems. The sociotechnical perspective of information behaviour, “how individuals encounter and make sense of their environment” (Case and Given, 2016, p. 4), is aligned with the neurosymbolic view making the full range of information behavior more applicable. With the concept of neurosymbolic machine learning still evolving, there is an opportunity for information behaviour research to help define it and by doing so advance human-centered artificial intelligence.

\section{Bridging the Divide}

The field of information behaviour has explored different paradigms (Hartel, 2019) and it continues to seek new ones (Tang et al., 2021). With its focus on machine learning, this paper may be seen as a return to the “physical paradigm” (Hartel, 2019) or the paradigm of “system oriented” studies (P. Vakkari, 1999) of earlier information behaviour research, especially those focused on intelligent agents and information retrieval (Ingwersen, 1992; Ingwersen and Järvelin, 2005). However, while this study is specific to machine learning systems, the larger argument is for an amplified focus on applied or actionable information behaviour: how information behaviour research can inform real-world applications.

In the machine learning context, applying information behaviour will require bridging terminological and conceptual divides to find common ground, publishing and presenting in the research fora of the machine learning community, and perhaps most importantly, participating in machine learning projects or collaborations. The author’s two-year association with the Vector Institute (http://vectorinstitute.ai), one of Canada’s three premier, federally funded, artificial intelligence research and development incubators, revealed an openness and appetite for this sort of engagement. 

Huvila et al. call for “a serious dialogue and a small dose of clarity in what individual IBP [information behaviour and practices] studies are aiming at and capable of doing in such terms that are actionable in ISD [information systems design]” (Huvila et al., 2021, p. 11). With respect to the machine learning community, that same dialogue and clarity is necessary to open itself to the needed influences from fields like information behaviour.

\section{Conclusion}

Gorichanaz and Venkatagiri discuss the “expanding circle” of information behaviour and its connection to, but “estrangement” from, human-computer interaction and propose “joining forces in pursuit of a shared mission” (Gorichanaz and Venkatagiri, 2021 p.9). While the intersection of information behaviour with machine learning is less than that with human-computer interaction, a shared mission is discernable. Information behavior research can bring to machine learning a more nuanced view of information and informing, a better understanding of information need and how that affects the communication among people and systems, guidance on the nature of context and how to operationalize that in models and systems, and insights into bias, misinformation, and marginalization. In addition, the general and context specific information behaviour models can provide frameworks or theoretical foundations for machine learning to leverage. As the neurosymbolic paradigm for machine learning evolves, the alignment with information behaviour is expected to be to even more relevant and productive.

To return to Bates and Thoreau, it appears that information behaviour and machine learning have much that is “important to communicate” and that “great haste” is warranted.

\bibliographystyle{apalike} 
\nocite{*}
\bibliography{ibml}

\begin{thebibliography}{}

\bibitem[Agarwal, 2015]{agarwal_towards_2015}
Agarwal, N.~K. (2015).
\newblock Towards a definition of serendipity in information behaviour.
\newblock {\em Information Research}, 20(3).

\bibitem[Agarwal, 2017]{agarwal_exploring_2017}
Agarwal, N.~K. (2017).
\newblock {\em Exploring context in information behavior: {Seeker}, situation,
  surroundings, and shared identities}.
\newblock Morgan \& Claypool, Williston, VT.

\bibitem[Bates, 1989]{bates_design_1989}
Bates, M.~J. (1989).
\newblock The design of browsing and berrypicking techniques for the online
  search interface.
\newblock {\em Online Review}, 13(5):407--424.

\bibitem[Bates, 1999]{bates_invisible_1999}
Bates, M.~J. (1999).
\newblock The invisible substrate of information science.
\newblock {\em Journal of the American Society for Information Science},
  50(12):1043--1050.
\newblock Publisher: John Wiley \& Sons, Inc.

\bibitem[Belkin, 1980]{belkin_anomalous_1980}
Belkin, N.~J. (1980).
\newblock Anomalous states of knowledge as a basis for information retrieval.
\newblock {\em Canadian Journal of Information Science}, 5:133--143.

\bibitem[Bender et~al., 2021]{bender_dangers_2021}
Bender, E.~M., Gebru, T., McMillan-Major, A., and Shmitchell, S. (2021).
\newblock On the dangers of stochastic parrots: {Can} language models be too
  big?
\newblock In {\em Proceedings of the 2021 {ACM} {Conference} on {Fairness},
  {Accountability}, and {Transparency}}, {FAccT} '21, pages 610--623, New York,
  NY, USA. Association for Computing Machinery.

\bibitem[Birhane et~al., 2021]{birhane_values_2021}
Birhane, A., Kalluri, P., Card, D., Agnew, W., Dotan, R., and Bao, M. (2021).
\newblock The values encoded in machine learning research.
\newblock {\em arXiv:2106.15590 [cs]}.
\newblock arXiv: 2106.15590.

\bibitem[Bommasani et~al., 2021]{bommasani_opportunities_2021}
Bommasani, R., Hudson, D.~A., Adeli, E., Altman, R., Arora, S., von Arx, S.,
  Bernstein, M.~S., Bohg, J., Bosselut, A., Brunskill, E., Brynjolfsson, E.,
  Buch, S., Card, D., Castellon, R., Chatterji, N., Chen, A., Creel, K., Davis,
  J.~Q., Demszky, D., Donahue, C., Doumbouya, M., Durmus, E., Ermon, S.,
  Etchemendy, J., Ethayarajh, K., Fei-Fei, L., Finn, C., Gale, T., Gillespie,
  L., Goel, K., Goodman, N., Grossman, S., Guha, N., Hashimoto, T., Henderson,
  P., Hewitt, J., Ho, D.~E., Hong, J., Hsu, K., Huang, J., Icard, T., Jain, S.,
  Jurafsky, D., Kalluri, P., Karamcheti, S., Keeling, G., Khani, F., Khattab,
  O., Kohd, P.~W., Krass, M., Krishna, R., Kuditipudi, R., Kumar, A., Ladhak,
  F., Lee, M., Lee, T., Leskovec, J., Levent, I., Li, X.~L., Li, X., Ma, T.,
  Malik, A., Manning, C.~D., Mirchandani, S., Mitchell, E., Munyikwa, Z., Nair,
  S., Narayan, A., Narayanan, D., Newman, B., Nie, A., Niebles, J.~C.,
  Nilforoshan, H., Nyarko, J., Ogut, G., Orr, L., Papadimitriou, I., Park,
  J.~S., Piech, C., Portelance, E., Potts, C., Raghunathan, A., Reich, R., Ren,
  H., Rong, F., Roohani, Y., Ruiz, C., Ryan, J., Ré, C., Sadigh, D., Sagawa,
  S., Santhanam, K., Shih, A., Srinivasan, K., Tamkin, A., Taori, R., Thomas,
  A.~W., Tramèr, F., Wang, R.~E., Wang, W., Wu, B., Wu, J., Wu, Y., Xie,
  S.~M., Yasunaga, M., You, J., Zaharia, M., Zhang, M., Zhang, T., Zhang, X.,
  Zhang, Y., Zheng, L., Zhou, K., and Liang, P. (2021).
\newblock On the opportunities and risks of foundation models.
\newblock {\em arXiv:2108.07258 [cs]}.
\newblock arXiv: 2108.07258.

\bibitem[Bowker and Star, 1999]{bowker_sorting_1999}
Bowker, G.~C. and Star, S.~L. (1999).
\newblock {\em Sorting things out}.
\newblock MIT Press, Cambridge, MA.

\bibitem[Brooks, 2019]{brooks_better_2019}
Brooks, R.~A. (2019).
\newblock A better lesson.
\newblock Key.

\bibitem[Brown et~al., 2020]{brown_language_2020}
Brown, T.~B., Mann, B., Ryder, N., Subbiah, M., Kaplan, J., Dhariwal, P.,
  Neelakantan, A., Shyam, P., Sastry, G., Askell, A., Agarwal, S.,
  Herbert-Voss, A., Krueger, G., Henighan, T., Child, R., Ramesh, A., Ziegler,
  D.~M., Wu, J., Winter, C., Hesse, C., Chen, M., Sigler, E., Litwin, M., Gray,
  S., Chess, B., Clark, J., Berner, C., McCandlish, S., Radford, A., Sutskever,
  I., and Amodei, D. (2020).
\newblock Language {Models} are {Few}-{Shot} {Learners}.
\newblock In {\em 34th {Conference} on {Neural} {Information} {Processing}
  {Systems} ({NeurIPS})}, Vancouver, BC.
\newblock arXiv: 2005.14165.

\bibitem[Burkov, 2019]{burkov_hundred-page_2019}
Burkov, A. (2019).
\newblock {\em The hundred-page machine learning book}.
\newblock Themlbook.

\bibitem[Case and Given, 2016]{case_looking_2016}
Case, D.~O. and Given, L.~M. (2016).
\newblock {\em Looking for information: {A} survey of research on information
  seeking, needs and behavior}.
\newblock Library and information science ({New} {York}, {N}.{Y}.). Emerald
  Group Publishing, Bingley, UK, 4th ed. edition.

\bibitem[Chang and Rice, 1993]{chang_browsing:_1993}
Chang, S.-J.~L. and Rice, R.~E. (1993).
\newblock Browsing: {A} multidimentional framework.
\newblock {\em Annual Review of Information Science and Technology},
  28:231--276.

\bibitem[Chatman, 1996]{chatman_impoverished_1996}
Chatman, E.~A. (1996).
\newblock The impoverished life-world of outsiders.
\newblock {\em Journal of the American Society for Information Science},
  47(3):193--206.
\newblock Place: Washington, D.C Publisher: Wiley Subscription Services, Inc, A
  Wiley Company.

\bibitem[Chatterjee and Shenoy, 2021]{chatterjee_model-agnostic_2021}
Chatterjee, S. and Shenoy, P. (2021).
\newblock Model-agnostic fits for understanding information seeking patterns in
  humans.
\newblock In {\em Proceeding of the {AAAI} {Conference} on {Artificial}
  {Intelligence}}, volume 35(1), pages 784--791.
\newblock arXiv: 2012.04858.

\bibitem[Chen et~al., 2021]{chen_evaluating_2021}
Chen, M., Tworek, J., Jun, H., Yuan, Q., Pinto, H. P. d.~O., Kaplan, J.,
  Edwards, H., Burda, Y., Joseph, N., Brockman, G., Ray, A., Puri, R., Krueger,
  G., Petrov, M., Khlaaf, H., Sastry, G., Mishkin, P., Chan, B., Gray, S.,
  Ryder, N., Pavlov, M., Power, A., Kaiser, L., Bavarian, M., Winter, C.,
  Tillet, P., Such, F.~P., Cummings, D., Plappert, M., Chantzis, F., Barnes,
  E., Herbert-Voss, A., Guss, W.~H., Nichol, A., Paino, A., Tezak, N., Tang,
  J., Babuschkin, I., Balaji, S., Jain, S., Saunders, W., Hesse, C., Carr,
  A.~N., Leike, J., Achiam, J., Misra, V., Morikawa, E., Radford, A., Knight,
  M., Brundage, M., Murati, M., Mayer, K., Welinder, P., McGrew, B., Amodei,
  D., McCandlish, S., Sutskever, I., and Zaremba, W. (2021).
\newblock Evaluating large language models trained on code.
\newblock {\em arXiv:2107.03374 [cs]}.
\newblock arXiv: 2107.03374.

\bibitem[Choo and Auster, 1993]{choo_environmental_1993}
Choo, C.~W. and Auster, E. (1993).
\newblock Environmental scanning: {Acquisitions} and use of information by
  managers.
\newblock {\em Annual Review of Information Science and Technology},
  28:279--314.

\bibitem[Cole, 2012]{cole_information_2012}
Cole, C. (2012).
\newblock {\em Information need: {A} theory connecting information search to
  knowledge formation}.
\newblock {ASIST} monograph series. American Society for Information Science
  and Technology, Medford, New Jersey.

\bibitem[Collins and Ghahramani, 2021]{collins_lamda_2021}
Collins, E. and Ghahramani, Z. (2021).
\newblock {LaMDA}: {Our} breakthrough conversation technology.

\bibitem[Dervin, 1976]{dervin_strategies_1976}
Dervin, B. (1976).
\newblock Strategies for dealing with human information needs: {Information} or
  communication?
\newblock {\em Journal of Broadcasting}, 20(3):323--333.

\bibitem[Dietterich, 2018]{dietterich_reflections_2018}
Dietterich, T.~G. (2018).
\newblock Reflections on innateness in machine learning.

\bibitem[Ellis et~al., 1993]{ellis_comparison_1993}
Ellis, D., Cox, D., and Hall, K. (1993).
\newblock A comparison of the information seeking patterns of researchers in
  the physical and social sciences.
\newblock {\em Journal of Documentation}, 49:356--3969.

\bibitem[Erdelez, 1997]{erdelez_information_1997}
Erdelez, Sanda, E. (1997).
\newblock Information encountering: {A} conceptual framework for accidental
  information discovery.
\newblock In Vakkari, R., Savolainen, R., and Dervin, B., editors, {\em
  Proceedings of the international conference on research in information needs,
  seeking and use in different contexts}, pages 412--421. Taylor Graham.

\bibitem[Eriksson-Backa et~al., 2021]{eriksson-backa_enablers_2021}
Eriksson-Backa, K., Hirvonen, N., Enwald, H., and Huvila, I. (2021).
\newblock Enablers for and barriers to using {My} {Kanta} – {A} focus group
  study of older adults’ perceptions of the {National} {Electronic} {Health}
  {Record} in {Finland}.
\newblock {\em Informatics for Health and Social Care}, 46(4):399--411.
\newblock Publisher: Taylor \& Francis.

\bibitem[Fidel and Green, 2004]{fidel_many_2004}
Fidel, R. and Green, M. (2004).
\newblock The many faces of accessibility: {Engineers}' perception of
  information sources.
\newblock {\em Information Processing \& Management}, 40(3):463--581.

\bibitem[Fisher et~al., 2005]{fisher_theories_2005}
Fisher, K.~E., Erdelez, S., and McKechnie, L., editors (2005).
\newblock {\em Theories of information behavior}.
\newblock {ASIST} monograph series. American Society for Information Science
  and Technology, Medford, NJ.

\bibitem[Foster, 2004]{foster_nonlinear_2004}
Foster, A. (2004).
\newblock A nonlinear model of information-seeking behavior.
\newblock {\em Journal of the American Society for Information Science and
  Technology}, 55(3):228--237.
\newblock Place: Hoboken Publisher: Wiley Subscription Services, Inc, A Wiley
  Company.

\bibitem[Gibson and Martin, 2019]{gibson_resituating_2019}
Gibson, A.~N. and Martin, J.~D. (2019).
\newblock Re‐situating information poverty: {Information} marginalization and
  parents of individuals with disabilities.
\newblock {\em Journal of the Association for Information Science and
  Technology}, 70(5):476--487.
\newblock Place: Hoboken, USA Publisher: John Wiley \& Sons, Inc.

\bibitem[Gitelman, 2013]{gitelman_raw_2013}
Gitelman, L. (2013).
\newblock {\em "{Raw} data" is an oxymoron}.
\newblock Infrastructures series. The MIT Press, Cambridge, Massachusetts.

\bibitem[Gorichanaz, 2020]{gorichanaz_information_2020}
Gorichanaz, T. (2020).
\newblock {\em Information experience in theory and design}.
\newblock Emerald Publishing.

\bibitem[Gorichanaz and Venkatagiri, 2021]{gorichanaz_expanding_2021}
Gorichanaz, T. and Venkatagiri, S. (2021).
\newblock The expanding circles of information behavior and human–computer
  interaction.
\newblock {\em Journal of librarianship and information science}, pages 1--15.

\bibitem[Greyson, 2018]{greyson_information_2018}
Greyson, D. (2018).
\newblock Information triangulation: {A} complex and agentic everyday
  information practice.
\newblock {\em Journal of the American Society for Information Science},
  69(7):869--878.

\bibitem[Guzman and Lewis, 2020]{guzman_artificial_2020}
Guzman, A.~L. and Lewis, S.~C. (2020).
\newblock Artificial intelligence and communication: {A} human–machine
  communication research agenda.
\newblock {\em New Media \& Society}, 22(1):70--86.

\bibitem[Hanlon and McLeod, 2020]{hanlon_human_2020}
Hanlon, S. and McLeod, J. (2020).
\newblock Human information behaviour in conversation: {Understanding} the
  influence of informal conversation on learning in a political party.
\newblock {\em Information Research}, 25(4).
\newblock Publisher: University of Borås.

\bibitem[Hartel, 2019]{hartel_turn_2019}
Hartel, J. (2019).
\newblock Turn, turn, turn.
\newblock {\em Information Research}, 24(4).
\newblock Key.

\bibitem[Hayes-Roth et~al., 1983]{hayes-roth_building_1983}
Hayes-Roth, F., Waterman, D.~A., and Lenat, D.~B., editors (1983).
\newblock {\em Building expert systems}.
\newblock Addison-Wesley, Reading, Mass.

\bibitem[Huvila et~al., 2021]{huvila_information_2021}
Huvila, I., Enwald, H., Eriksson‐Backa, K., Liu, Y.-H., and Hirvonen, N.
  (2021).
\newblock Information behavior and practices research informing information
  systems design.
\newblock {\em Journal of the Association for Information Science and
  Technology}, pages 1--15.

\bibitem[Ingwersen, 1992]{ingwersen_information_1992}
Ingwersen, P. (1992).
\newblock {\em Information retrieval interaction}.
\newblock Taylor Graham, London.

\bibitem[Ingwersen and Järvelin, 2005]{ingwersen_turn:_2005}
Ingwersen, P. and Järvelin, K. (2005).
\newblock {\em The turn: {Integration} of information seeking and retrieval in
  context}.
\newblock Springer, Dordrecht.

\bibitem[Julien and O'Brien, 2014]{julien_information_2014}
Julien, H. and O'Brien, M. (2014).
\newblock Information behaviour research: {Where} have we been, where are we
  going?
\newblock {\em Canadian Journal of Information and Library Science},
  38(4):239--250.

\bibitem[Karlova and Fisher, 2013]{karlova_social_2013}
Karlova, N. and Fisher, K. (2013).
\newblock A social diffusion model of misinformation and disinformation for
  understanding human information behaviour.
\newblock {\em Information Research}, 18(1).
\newblock Publisher: Professor T.D. Wilson.

\bibitem[Kuhlthau, 1991]{kuhlthau_inside_1991}
Kuhlthau, C.~C. (1991).
\newblock Inside the search process: {Information} seeking from the user's
  perspective.
\newblock {\em Journal of the American Society for Information Science},
  42(5):361--71.

\bibitem[Lacoste et~al., 2021]{lacoste_toward_2021}
Lacoste, A., Sherwin, E.~D., Kerner, H., Alemohammad, H., Lütjens, B., Irvin,
  J., Dao, D., Chang, A., Gunturkun, M., Drouin, A., Rodriguez, P., and
  Vazquez, D. (2021).
\newblock Toward foundation models for earth monitoring: {Proposal} for a
  climate change benchmark.
\newblock {\em arXiv:2112.00570 [physics]}.
\newblock arXiv: 2112.00570.

\bibitem[Leckie et~al., 1996]{leckie_modeling_1996}
Leckie, G.~J., Pettigrew, K.~E., and Sylvain, C. (1996).
\newblock Modeling the information seeking of professionals: {A} general model
  derived from research on engineers, health care professionals, and lawyers.
\newblock {\em Library Quarterly}, 66(2):161--93.

\bibitem[LeCun et~al., 2015]{lecun_deep_2015}
LeCun, Y., Bengio, Y., and Hinton, G. (2015).
\newblock Deep learning.
\newblock {\em Nature}, 521:436--444.

\bibitem[Lee et~al., 2021]{lee_information_2021}
Lee, L., Ocepek, M.~G., and Makri, S. (2021).
\newblock Information behavior patterns: {A} new theoretical perspective from
  an empirical study of naturalistic information acquisition.
\newblock {\em Journal of the Association for Information Science and
  Technology}.
\newblock \_eprint:
  https://asistdl.onlinelibrary.wiley.com/doi/pdf/10.1002/asi.24595.

\bibitem[Littman et~al., 2021]{littman_gathering_2021}
Littman, M.~L., Ajunwa, I., Berger, G., Boutilier, G., Currie, M., Doshi-Velez,
  F., and Hadfield, G. (2021).
\newblock Gathering strength, gathering storms: {The} one hundred year study on
  artificial intelligence ({AI100}) 2021 study panel report.
\newblock Technical report, Stanford University.

\bibitem[Liu, 2017]{liu_toward_2017}
Liu, J. (2017).
\newblock Toward a unified model of human information behavior: {An}
  equilibrium perspective.
\newblock {\em Journal of Documentation}, 73(4):666--688.

\bibitem[Ma and Lund, 2021]{ma_evolution_2021}
Ma, J. and Lund, B. (2021).
\newblock The evolution and shift of research topics and methods in library and
  information science.
\newblock {\em Journal Of the American Society for Information Science},
  72:1059--1073.

\bibitem[Makri, 2020]{makri_information_2020}
Makri, S. (2020).
\newblock Information informing design: {Information} science research with
  implications for the design of digital information environments.
\newblock {\em Journal of the Association for Information Science and
  Technology}, 71(11):1402--1412.
\newblock \_eprint:
  https://asistdl.onlinelibrary.wiley.com/doi/pdf/10.1002/asi.24418.

\bibitem[Marcus, 2018]{marcus_deep_2018}
Marcus, G. (2018).
\newblock Deep learning: {A} critical appraisal.
\newblock {\em arXiv:1801.00631 [cs, stat]}.
\newblock arXiv: 1801.00631.

\bibitem[Marcus, 2020]{marcus_next_2020}
Marcus, G. (2020).
\newblock The next decade in {AI}: {Four} steps towards robust artificial
  intelligence.
\newblock {\em arXiv:2002.06177 [cs]}.
\newblock arXiv: 2002.06177.

\bibitem[Marcus and Davis, 2019]{marcus_rebooting_2019}
Marcus, G. and Davis, E. (2019).
\newblock {\em Rebooting {AI}: {Building} artificial intelligence we can
  trust}.
\newblock Pantheon, New York.

\bibitem[Marcus and Davis, 2021]{marcus_has_2021}
Marcus, G. and Davis, E. (2021).
\newblock Has {AI} found a new foundation?
\newblock {\em The Gradient}.

\bibitem[McCay-Peet and Toms, 2018]{mccay-peet_researching_2018}
McCay-Peet, L. and Toms, E. (2018).
\newblock {\em Researching serendipity in digital information environments}.
\newblock Synthesis lectures on information concepts, retrieval, and services ;
  \#59. Morgan \& Claypool, San Rafael, CA.

\bibitem[Nayak, 2021]{nayak_mum_2021}
Nayak, P. (2021).
\newblock {MUM}: {A} new {AI} milestone for understanding information.

\bibitem[Nikolenko, 2019]{nikolenko_synthetic_2019}
Nikolenko, S.~I. (2019).
\newblock Synthetic data for deep learning.
\newblock {\em arXiv:1909.11512 [cs]}.
\newblock arXiv: 1909.11512.

\bibitem[Ocepek, 2018]{ocepek_bringing_2018}
Ocepek, M.~G. (2018).
\newblock Bringing out the everyday in everyday information behavior.
\newblock {\em Journal of Documentation}, 74(2):398--411.

\bibitem[Paullada et~al., 2020]{paullada_data_2020}
Paullada, A., Raji, I.~D., Bender, E.~M., Denton, E., and Hanna, A. (2020).
\newblock Data and its (dis)contents: {A} survey of dataset development and use
  in machine learning research.
\newblock {\em arXiv:2012.05345 [cs]}.
\newblock arXiv: 2012.05345.

\bibitem[Rahwan et~al., 2019]{rahwan_machine_2019}
Rahwan, I., Cebrian, M., Obradovich, N., Bongard, J., Bonnefon, J.-F.,
  Breazeal, C., Crandall, J.~W., Christakis, N.~A., Couzin, I.~D., Jackson,
  M.~O., Jennings, N.~R., Kamar, E., Kloumann, I.~M., Larochelle, H., Lazer,
  D., McElreath, R., Mislove, A., Parkes, D.~C., Pentland, A., Roberts, M.~E.,
  Shariff, A., Tenenbaum, J.~B., and Wellman, M. (2019).
\newblock Machine behaviour.
\newblock {\em Nature}, 568(7753):477--486.
\newblock Key.

\bibitem[Ridley, 2019]{ridley_autonomous_2019}
Ridley, M. (2019).
\newblock The autonomous turn in information behaviour.
\newblock {\em Information Research}, 24(1).

\bibitem[Ridley, 2022]{ridley_machine_2022}
Ridley, M. (2022).
\newblock Machine information behaviour.
\newblock In Hervieux, S. and Wheatley, A., editors, {\em The rise of {AI}:
  {Implications} and applications of artificial intelligence in academic
  libraries}, pages 175--188. Association of College and University Libraries.

\bibitem[Riedl, 2019]{riedl_human-centered_2019}
Riedl, M.~O. (2019).
\newblock Human-centered artificial intelligence and machine learning.
\newblock {\em Human Behavior and Emerging Technologies}, 1(1):33--36.
\newblock \_eprint: https://onlinelibrary.wiley.com/doi/pdf/10.1002/hbe2.117.

\bibitem[Rothchild et~al., 2021]{rothchild_c5t5_2021}
Rothchild, D., Tamkin, A., Yu, J., Misra, U., and Gonzalez, J. (2021).
\newblock {C5T5}: {Controllable} generation of organic molecules with
  transformers.
\newblock {\em arXiv:2108.10307 [cs]}.
\newblock arXiv: 2108.10307.

\bibitem[Rubin, 2019]{rubin_disinformation_2019}
Rubin, V.~L. (2019).
\newblock Disinformation and misinformation triangle.
\newblock {\em Journal of Documentation}, 75(5):1013--1034.

\bibitem[Russell, 2019]{russell_human_2019}
Russell, S. (2019).
\newblock {\em Human compatible: {Artificial} intelligence and the problem of
  control}.
\newblock Viking, New York.

\bibitem[Russell, 2021]{russell_history_2021}
Russell, S. (2021).
\newblock The history and future of {AI}.
\newblock {\em Oxford Review of Economic Policy}, 37(3):509--520.

\bibitem[Sandstrom, 1994]{sandstrom_optimal_1994}
Sandstrom, P.~E. (1994).
\newblock An optimal foraging approach to information seeking and use.
\newblock {\em The Library Quarterly}, 64(4):414--449.

\bibitem[Savolainen, 2008]{savolainen_everyday_2008}
Savolainen, R. (2008).
\newblock {\em Everyday information practices: {A} social phenomenological
  perspective}.
\newblock Scarecrow Press, Lanham, Md.

\bibitem[Savolainen and Thomson, 2022]{savolainen_assessing_2022}
Savolainen, R. and Thomson, L. (2022).
\newblock Assessing the theoretical potential of an expanded model for everyday
  information practices.
\newblock {\em Journal of the Association for Information Science and
  Technology}, 73(4):511--527.
\newblock \_eprint:
  https://asistdl.onlinelibrary.wiley.com/doi/pdf/10.1002/asi.24589.

\bibitem[Seaver, 2021]{seaver_seeing_2021}
Seaver, N. (2021).
\newblock Seeing like an infrastructure: {Avidity} and difference in
  algorithmic recommendation.
\newblock {\em Cultural Studies}, 35(4-5):771--791.
\newblock Publisher: Routledge \_eprint:
  https://doi.org/10.1080/09502386.2021.1895248.

\bibitem[Steinhardt, 2021]{steinhardt_risks_2021}
Steinhardt, J. (2021).
\newblock On the risks of emergent behavior in foundation models.

\bibitem[Sutton, 2019]{sutton_bitter_2019}
Sutton, R.~S. (2019).
\newblock The bitter lesson.
\newblock Key.

\bibitem[Tamkin et~al., 2021]{tamkin_understanding_2021}
Tamkin, A., Brundage, M., Clark, J., and Ganguli, D. (2021).
\newblock Understanding the capabilities, limitations, and societal impact of
  large language models.
\newblock {\em arXiv:2102.02503 [cs]}.
\newblock arXiv: 2102.02503.

\bibitem[Tang et~al., 2021]{tang_framing_2021}
Tang, R., Mehra, B., Du, J.~T., and Zhao, Y.~C. (2021).
\newblock Framing a discussion on paradigm shift(s) in the field of
  information.
\newblock {\em Journal of the Association for Information Science and
  Technology}, 72(2):253--258.
\newblock \_eprint: https://onlinelibrary.wiley.com/doi/pdf/10.1002/asi.24404.

\bibitem[Taylor, 2018]{taylor_improving_2018}
Taylor, M.~E. (2018).
\newblock Improving reinforcement learning with human input.
\newblock In {\em Proceedings of the {Twenty}-{Seventh} {International} {Joint}
  {Conference} on {Artificial} {Intelligence} ({IJCAI}-18)}, pages 5724--5728,
  Stockholm.

\bibitem[Taylor, 1968]{taylor_question-negotiation_1968}
Taylor, R.~S. (1968).
\newblock Question-negotiation and information seeking in libraries.
\newblock {\em College \& Research Libraries}, 29(3):178--194.

\bibitem[Thoreau, 1854]{thoreau_walden_1854}
Thoreau, H.~D. (1854).
\newblock {\em Walden: {A} fully annotated edition}.
\newblock Yale University Press, New Haven.

\bibitem[Vakkari, 1999]{vakkari_task_1999}
Vakkari, P. (1999).
\newblock Task complexity, problem structure and information actions:
  {Integrating} studies on information seeking and retrieval.
\newblock {\em Information processing \& management}, 35(6):819--837.
\newblock Publisher: Elsevier Ltd.

\bibitem[Vanhoucke, 2021]{vanhoucke_three_2021}
Vanhoucke, V. (2021).
\newblock Three grand challenges in machine learning.

\bibitem[Weidinger et~al., 2021]{weidinger_ethical_2021}
Weidinger, L., Mellor, J., Rauh, M., Griffin, C., Uesato, J., Huang, P.-S.,
  Cheng, M., Glaese, M., Balle, B., Kasirzadeh, A., Kenton, Z., Brown, S.,
  Hawkins, W., Stepleton, T., Biles, C., Birhane, A., Haas, J., Rimell, L.,
  Hendricks, L.~A., Isaac, W., Legassick, S., Irving, G., and Gabriel, I.
  (2021).
\newblock Ethical and social risks of harm from language models.
\newblock Technical report, DeepMind.

\bibitem[Willson et~al., 2021]{willson_retrospective_2021}
Willson, R., Julien, H., and Allen, D. (2021).
\newblock Retrospective special issue - {Information} behaviour.
\newblock {\em Journal of the Association for Information Science and
  Technology}.
\newblock \_eprint:
  https://asistdl.onlinelibrary.wiley.com/doi/pdf/10.1002/asi.24557.

\bibitem[Wilson, 2016]{wilson_general_2016}
Wilson, T.~D. (2016).
\newblock A general theory of human information behaviour.
\newblock {\em Information Research}, 21(4).
\newblock Key.

\bibitem[Wilson, 2018]{wilson_diffusion_2018}
Wilson, T.~D. (2018).
\newblock The diffusion of information behaviour research across disciplines.
\newblock {\em Information Research}, 23(4).
\newblock Publisher: University of Borås.

\bibitem[Wilson, 2020a]{wilson_exploring_2020}
Wilson, T.~D. (2020a).
\newblock {\em Exploring information behaviour: {An} introduction}.
\newblock Information Research.
\newblock Key.

\bibitem[Wilson, 2020b]{wilson_transfer_2020}
Wilson, T.~D. (2020b).
\newblock The transfer of theories and models from information behaviour
  research into other disciplines.
\newblock {\em Information Research}, 25(3).

\bibitem[Yuan et~al., 2021]{yuan_florence_2021}
Yuan, L., Chen, D., Chen, Y.-L., Codella, N., Dai, X., Gao, J., Hu, H., Huang,
  X., Li, B., Li, C., Liu, C., Liu, M., Liu, Z., Lu, Y., Shi, Y., Wang, L.,
  Wang, J., Xiao, B., Xiao, Z., Yang, J., Zeng, M., Zhou, L., and Zhang, P.
  (2021).
\newblock Florence: {A} new foundation model for computer vision.
\newblock {\em arXiv:2111.11432 [cs]}.
\newblock arXiv: 2111.11432 version: 1.

\bibitem[Zhifei and Meng~Joo, 2012]{gao_survey_2012}
Zhifei, S. and Meng~Joo, E. (2012).
\newblock A survey of inverse reinforcement learning techniques.
\newblock {\em International Journal of Intelligent Computing and Cybernetics},
  5(3):293--311.

\end{thebibliography}

\end{document}